\documentclass[letterpaper, 10 pt, journal, twoside]{IEEEtran}

\IEEEoverridecommandlockouts
\usepackage{amsmath, amssymb}

\usepackage{graphics} 
\usepackage{epsfig} 
\usepackage{times} 
\usepackage{amsmath} 
\usepackage{amssymb}  
\usepackage[table]{xcolor}
\usepackage[ruled,linesnumbered, noend]{algorithm2e}
\SetKwRepeat{Do}{do}{while}
\usepackage{adjustbox}
\usepackage{wrapfig}

\newcommand{\ft}{F$\backslash$T }
\newcommand{\se}{AllSight }
\newcommand{\sep}{AllSight}

\usepackage{soul}
\usepackage{booktabs}

\usepackage{cite}
\usepackage{comment}
\usepackage{multirow}
\usepackage{graphicx}
\usepackage{wrapfig}
\usepackage{balance}
\usepackage{subfigure}
\usepackage{capt-of}

\newcommand{\ve}[1]{\mathbf{#1}} 
\newcommand{\tve}[1]{\tilde{\mathbf{#1}}} 

\usepackage{color, colortbl}
\definecolor{Gray}{gray}{0.9}

\usepackage{pifont}

\title{AllSight: A Low-Cost and High-Resolution Round Tactile Sensor with Zero-Shot Learning Capability}

\author{Osher Azulay, Nimrod Curtis, Rotem Sokolovsky, Guy Levitski, \\Daniel Slomovik, Guy Lilling and Avishai Sintov
\thanks{O. Azulay, N. Curtis, R. Sokolovsky, G. Levistky, D. Slomovik, G. Lilling and  A. Sintov are with the School of Mechanical Engineering, Tel-Aviv University, Israel. Corresponding Author: osherazulay@mail.tau.ac.il.
}
}

\begin{document}


\maketitle

\begin{abstract}
Tactile sensing is a necessary capability for a robotic hand to perform fine manipulations and interact with the environment. Optical sensors are a promising solution for high-resolution contact estimation. Nevertheless, they are usually not easy to fabricate and require individual calibration in order to acquire sufficient accuracy. In this letter, we propose \textit{\sep}, an optical tactile sensor with a round 3D structure designed for robotic in-hand manipulation tasks. \se is mostly 3D printed including a novel and simplified fabrication processes. This makes it low-cost, modular, durable and in the size of a human thumb while with a large contact surface. We show the ability of \se to learn and estimate a full contact state, i.e., contact position, forces and torsion. With that, an experimental benchmark between various configurations of illumination and contact elastomers are provided. Furthermore, the robust design of \se provides it with a unique zero-shot capability such that a practitioner can fabricate the open-source design and have a ready-to-use state estimation model. A set of experiments demonstrates the accurate state estimation performance of \sep.

\end{abstract}


\section{Introduction}
\label{sec:introduction}

\IEEEPARstart{T}{he} sense of touch endows humans with neural sensory-motor feedback regarding the shape, weight and texture of objects within contact \cite{Abraira2013}. Hence, touch is vital for humans in order to ensure stable grasps and safe object manipulations \cite{Johansson2009}. 
Similar to humans, robots require touch sensing in order to acquire information regarding the state of contact events. In order to manipulate objects effectively in complex and changing environments, a robot must be able to perceive when, where and how it is interacting with the objects \cite{Dikhale2022,flx2022finger}. Touch, or tactile sensing, can augment visual perception or replace it when occlusions occur by the robot fingers themselves or by obstacles. It has the potential to enable robots to infer about the object's relative state, geometry and texture \cite{Abad2020}. Accurate and low-cost high-resolution tactile sensors that can provide a full state of contact, namely contact locations and forces, would have a significant role in, for example, material handling, assembly \cite{Tang2016}, in-hand manipulation \cite{Kappassov2015} and prosthesis \cite{Wu2018}.



\begin{figure}[t]
\centering
\includegraphics[width=0.8\linewidth]{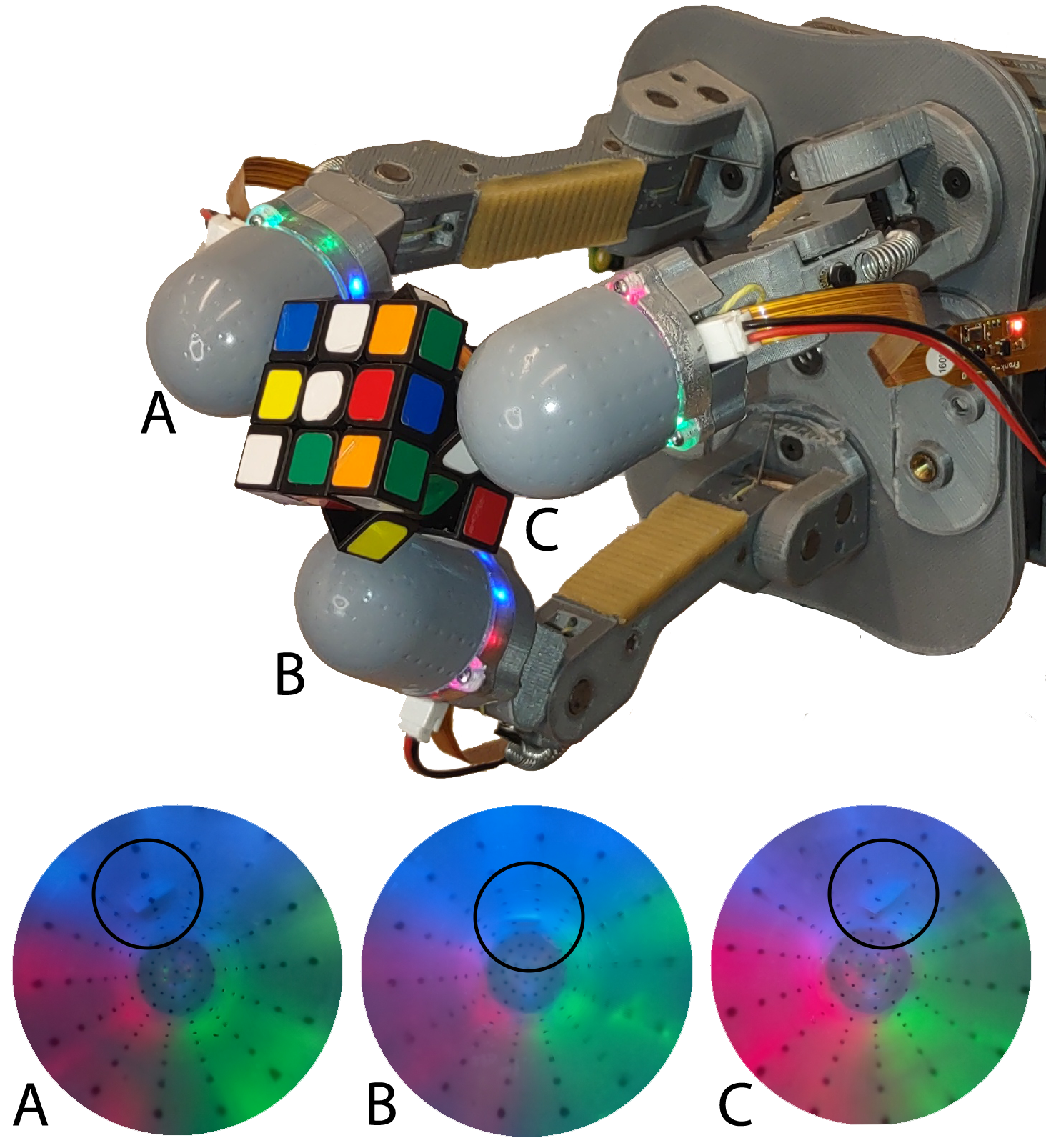} 
\caption{Three \textit{\se} sensors on the fingers of an OpenHand Model-O \cite{Ma2017YaleOP} robotic hand. The sensors provide real-time tactile images for contact state estimations during the manipulation of an object. Surface deformations due to contact are marked with a circle at the bottom tactile images.}
\label{fig:intro}
\vspace{-0.7cm}
\end{figure}

Tactile sensors are commonly used to measure a range of touch stimuli including contact pressure, vibrations, deformation of the contact pad and surface texture \cite{Liu2020}. 
Within the range of tactile usages, a variety of tactile sensing technologies exists including force sensitive resistors, capacitive transducers, photoelectric sensing and piezo-resistors \cite{Nguyen2022}.
Although they can provide useful data, these sensors are usually designed for specific tasks in limited environments. 
Recently, camera-based optical tactile sensors have become increasingly common due to high-resolution signals and soft contact surfaces \cite{taylor2022gelslim, lambeta2020digit}. An optical sensor typically uses an internal camera to track the deformation of a soft elastomer upon contact with an object \cite{ward2018tactip}. A captured image can encode information regarding the state of contact, i.e., contact location with respect to the sensor's coordinate frame and contact forces. Despite the abundance in configurations of optical sensors, they yet to provide a robust tactile solution and are limited in various aspects. 
\begin{figure*}[t]
\centering
\includegraphics[width=\linewidth,keepaspectratio]{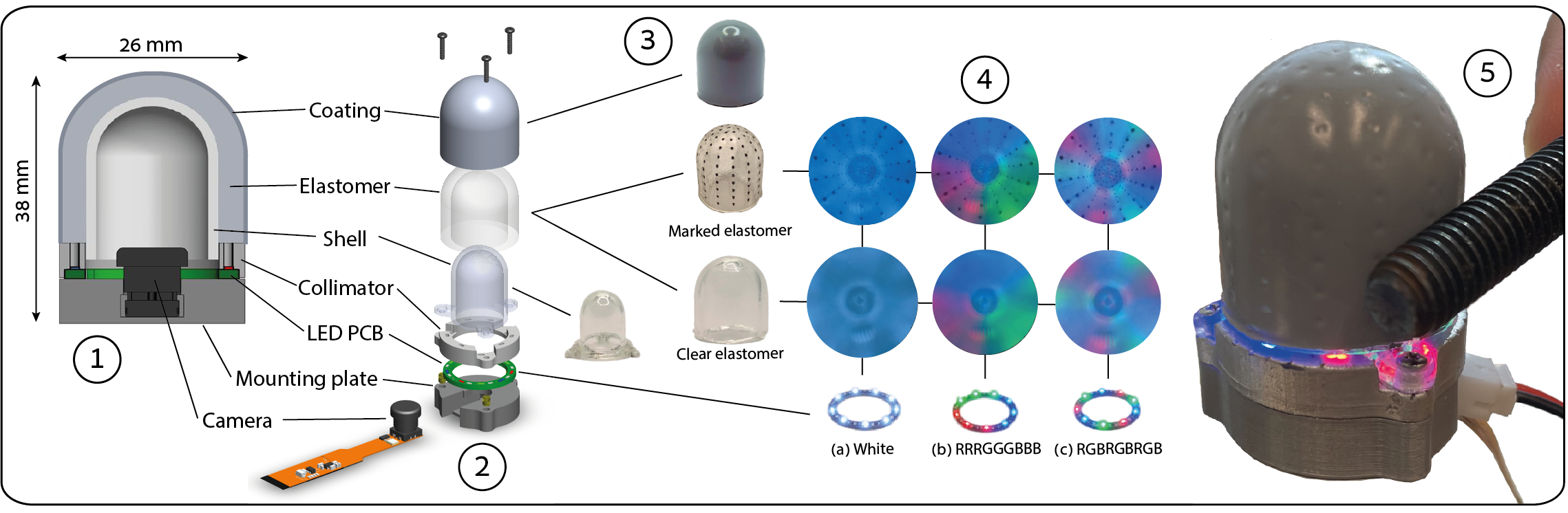} 
\vspace{-0.6cm}
\caption{Illustration of \se (1) assembled and (2) in an exploded view. (3) Images of the corresponding fabricated parts are seen including marked and clear elastomers. (4) Three different LED illumination configurations when (5) \se is in contact with a screw: (4a) white, (4b) RRRGGGBBB and (4c) RGBRGBRGB. Top and bottom rows of the camera view show elastomers with and without markers, respectively.}
\label{fig:intro_banner}
\vspace{-0.5cm}
\end{figure*}

While camera-based sensors are not a new notion, significant advancement and integration in robotic systems were achieved in the last few years with the increase of computing capabilities and hardware minimization. Various small sensors with flat contact surfaces were introduced \cite{yuan2017gelsight,lambeta2020digit,taylor2022gelslim,wang2022tacto}. However, these may have difficulties in general manipulation tasks due to the flat contact surfaces. Hence, other sensors introduced spatial surface geometries \cite{ward2018tactip,padmanabha2020omnitact}. Nevertheless, these sensors yet to provide complete and reliable contact information. Some were not demonstrated to provide a complete contact state \cite{romero2020soft}, are limited in load forces \cite{xu2023efficient} or require complex and expensive fabrication process \cite{sun2022soft}. Moreover, no sensor has demonstrated an ability for zero-shot transfer of a trained contact state estimation model to a new one. 
A comparative summary of 3D optical-based tactile sensors is given in Table \ref{tb:sota} while mentioning whether various features were demonstrated.

In this paper, we cope with the limitations of previous optical designs and present a novel spatial sensor termed \textit{\sep}. 
\se is designed to be small and low-cost for the use on multi-finger hands in in-hand manipulation tasks. The 3D contact surface of \se is in the shape of a cylinder with an hemispherical end as seen in Figure \ref{fig:intro}. Such a round structure expands manipulation capabilities where, unlike with flat sensors, finger rolling on the surface of the object is possible. Unlike prior work, most of the sensor's components, excluding electronics and elastomer, are printable. In a novel process, the transparent shell of \se is 3D printed making the sensor low-cost, easily fabricated and more accessible. Similarly, unlike prior approaches that require special tools for fabricating markers on the elastomer, an easy 3D printing process  is proposed. 

While fabricated in a low-cost process, \se is shown to provide an accurate full contact state including position, normal and tangential forces, and torsion. The sensor is the smallest of its kind while able to measure larger forces than previous designs and up to 15~N. In addition, the fabrication process results in a durable sensor able to withstand high and recurring loads. \se is modular with easily replaceable components where different types of elastomer and illumination can be rapidly replaced. A practitioner can choose to use an elastomer with or without markers based on sensing requirements. Through a comparative analysis of \sep, we try to answer some fundamental questions in the design of optical tactile sensors such as preferred illumination and surface texture. 
Structure and various tested sensor configurations are seen in Figure \ref{fig:intro_banner}.

The design, trained models, simulation environment, collected datasets, and code are provided open-source\footnote{\se Open-source design, fabrication instructions, trained models, code and simulation: \texttt{https://github.com/osheraz/allsight}} for the benefit of the community and to advance research in the field. However, the trained models should provide sufficient accuracy on a newly fabricated sensor. Therefore, we analyze the transfer learning capabilities of \se in zero-shot and in fine-tuning with limited new data. We show that these can be done by pre-training with real data collected from a source sensor or with simulated data from TACTO, a physics engine tactile simulation \cite{wang2022tacto}. Overall, \se is capable of achieving sufficient accuracy in zero-shot transfer and high accuracy with fine-tuning on limited new data. While not as low as the accuracy with the source sensor, zero-shot inference on a new sensor is shown to be able to provide sufficient accuracy for general tasks. Consequently, advanced and novice practitioners can have access to low-cost, easy to fabricate and reproducible sensors with a ready-to-use trained model.

A prominent goal of this paper is to share insights with the robotics community and to help overcome the numerous bottlenecks faced in the fabrication of spatial tactile sensors. We aim to encourage the wider adoption and development of such sensing technology. To summarize, the contributions of this work are as follow. First, we propose a novel design of spatial optical tactile sensor termed \se which is compact and with high-resolution. Since most of the parts are 3D printed including the transparent shell, \se is low-cost, easy to fabricate, modular and available open-source. 
An informative comparative analysis is provided involving various known sensor configurations which could assist practitioners in design choices. Finally, we exhibit the unique ability of \se to transfer to newly fabricated sensors through zero-shot and fine-tuning. To the best of the author’s knowledge, \se is the only 3D optical-based sensor that measures the entire contact state, capable of zero-shot learning
and is available open-source.

\section{Related work}
\label{sec:related_work}
\begin{table*}[ht]
\centering
\caption{\small State-of-the-art comparison of optical-based tactile sensors with a 3D contact surface}
\label{tb:sota}
\vspace{-0.3cm}
\begin{adjustbox}{width=0.9\textwidth}
\begin{tabular}{lcc ccc c ccccc ccc}
\hline
\multirow{2}{*}{Sensor} &&& \multicolumn{3}{c}{Transduction Method}   
&& \multicolumn{5}{c}{Contact state features$^*$} & Max. load & \multirow{2}{*}{Dimension (mm)} & \multirow{2}{*}{Weight (g)}   \\ \cline{4-6}\cline{8-12}
 &&& Camera & PS & Markers 
 && $\ve{x}$ & $f_z$ & $f_x,f_y$ & $\tau$ & $d$ & $f_z$ (N) & & \\ 
\hline

\rowcolor{Gray}
TacTip & 2018 & \cite{ward2018tactip} & \checkmark & & \checkmark &&  \checkmark &&&& & - & $40\times40\times85$ & - \\ 

Romero et al. & 2020 & \cite{romero2020soft} & \checkmark & \checkmark &  & &&&&& \checkmark & - & $28\times30\times35$ & - \\ 

\rowcolor{Gray}
GelTip & 2020 & \cite{gomes2020geltip} & \checkmark & \checkmark &   && \checkmark &&&& & - & $30\times30\times100$ & -\\ 

Omnitact & 2020 & \cite{padmanabha2020omnitact} & $5\times$\checkmark & \checkmark &  && \checkmark &&&& & - & $30\times30\times33$ & - \\

\rowcolor{Gray}
Insight & 2022 & \cite{sun2022soft} & \checkmark & \checkmark &   && \checkmark & \checkmark & \checkmark & && 2   & $40\times40\times 70$ & - \\ 

DenseTact & 2023 & \cite{do2023densetact}  & \checkmark & \checkmark & rand. & &  & \checkmark & \checkmark & \checkmark & \checkmark & 11 & $32\times32\times43$ & 34 \\ 

\rowcolor{Gray}
GelSight360 & 2023 & \cite{Tippur2023} & \checkmark & \checkmark &    && &&&& \checkmark & - & $28\times28\times50$ & - \\

\hline
\se && Ours & \checkmark & \checkmark & \checkmark && \checkmark & \checkmark & \checkmark & \checkmark & \checkmark & 15 & $26\times28\times38$ & 15 \\ \hline

\multicolumn{14}{l}{\footnotesize $^*$$\ve{x}$ denotes contact position; $f_z$ denotes normal force; $f_x$ and $f_y$ denote tangential forces; and $\tau$ denotes torsion about the normal at the } \\
\multicolumn{14}{l}{\footnotesize  contact surface; $d$ denotes penetration depth.}\\
\end{tabular}
\end{adjustbox}
\vspace{-0.7cm}
\end{table*}

Seminal work on optical sensors introduced the use of a black and white camera for observing the deformation of a soft membrane through a glass plate \cite{Begej1988}. Later work have shown the ability to use the same technology for a round or finger-shaped sensor \cite{Ferrier2000,Chorley2009}. The GelSight is the first to present a relatively small tactile finger-tip with a flat pad able to measure high-resolution contact geometry \cite{yuan2017gelsight}. Photometric Stereo (PS) was integrated where surface normals during contact deformation are estimated by observing the object under different lighting conditions. Hence, contact force, slip and shape were inferred by observing deformation and calculating geometry gradients. While exhibiting good performance, GelSight and similar ones (e.g., \cite{taylor2022gelslim,wang2022tacto}) may have difficulties in general dexterous manipulation tasks due to the flat contact surface. Flat sensors require constant alignment with the surface of the object and may not maintain contact during object sliding and rolling \cite{romero2020soft}. Hence, the TacTip set of sensors was introduced having a variety of different contact pads including flat and hemi-spherical ones \cite{ward2018tactip}. However, the contact pads did not include a rigid support for the elastomer and, thus, were reported to be too soft for feasible manipulation tasks \cite{xu2023efficient}. 

Tactile sensors that can efficiently manipulate objects must have a spatial surface structure. Yet, recent attempts to develop tactile sensors with 3D sensing surfaces have raised a number of challenges. To begin with, creating tactile sensors with round contact surfaces 
can be difficult from a manufacturing standpoint. Fabrication may require intricate designs and the use of high-budget machinery such as industrial (e.g., Stratasys) \cite{ward2018tactip} or Aluminum 3D printers \cite{sun2022soft}. Furthermore, the outer layer of the sensor is in constant contact with objects during use and, therefore, the surface pad is prone to wear and tear over time, adding to the complexity of creating a round tactile sensor that is both sensitive and reliable \cite{piacenza2020sensorized}.
It can also be challenging to make the sensor modular for convenient component replacement and easy-to-use through a plug-and-play interface \cite{lambeta2020digit}.

Researchers have conducted extensive application studies with flat sensors regarding contact localization \cite{lambeta2020digit}, depth reconstruction \cite{wang2021gelsight, yuan2017gelsight} and 
directional force distribution \cite{taylor2022gelslim}.
Sensors with round contact surface, on the other hand, yet to provide complete and reliable contact information. Some sensors can only provide contact localization \cite{gomes2020geltip, padmanabha2020omnitact, romero2020soft} or shape reconstruction \cite{Tippur2023} with no load information . However, manipulation capabilities require also information regarding contact forces. Recent sensor developments have tried to provide full contact state. A cone-shape thumb-sized sensor, for instance, provides a full force map along with contact localization \cite{sun2022soft}. However and due to its skeletal structure, it is sensitive to object penetration and, hence, limited to contact forces of up to 2~N. Similarly, the DenseTact provides contact loads through an hemi-spherical pad. A randomized pattern was added to the surface of the contact pad in order to increase features in the images \cite{do2023densetact}. The ability for transfer learning was also demonstrated in a limited setting without zero-shot and with some portion of new data used for calibrating the target sensor. The hemi-sphere of DenseTact is made of an elastomer without a rigid structure. This and the lack of a cylindrical extension to the hemi-sphere may reduce its durability and applicability in manipulation tasks. Transfer learning was also recently demonstrated in classification on the DIGIT flat sensor \cite{Higuera2023}. In the work, a diffusion model was trained to generate realistic tactile images and later calibrated to unseen sensors. 
In conclusion, prior work on optical tactile sensors lacks 3D contact surfaces that are simple and low-cost to fabricate, accessible through an open-source design, and can enable zero-shot model learning.

\section{Design and Fabrication}
\label{sec:design}
\begin{figure*}[ht!]
\centering
\includegraphics[width=\textwidth,keepaspectratio]{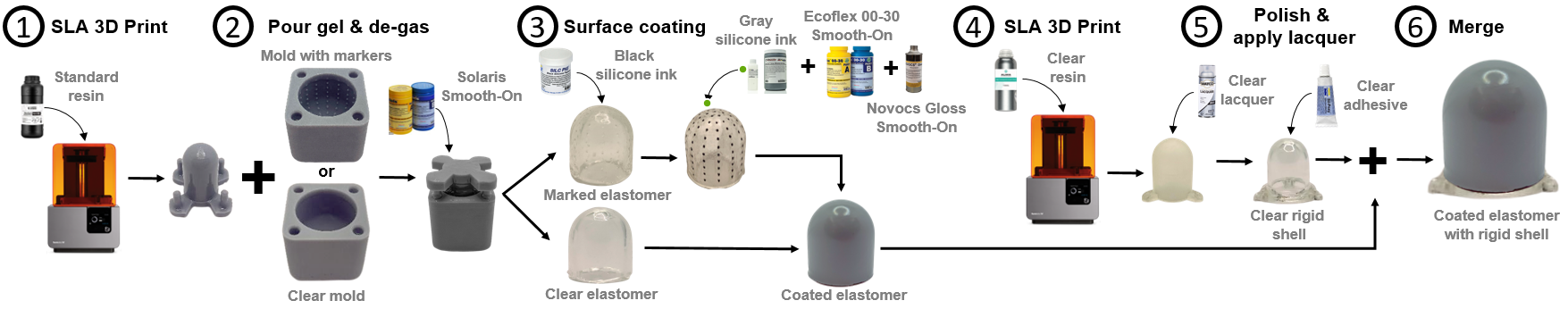}
\vspace{-0.7cm}
\caption{Steps 1-3 depict the fabrication process of the elastomer including (1) mold printing, (2) molding and (3) reflective coating. Steps 4-5 show the fabrication of the clear rigid shell through (4) 3D printing with clear resin and (5) polishing. In step 6, the coated elastomer is glued onto the rigid shell.}
\label{fig:Manufacturing}
\vspace{-0.5cm}
\end{figure*}

\se is an optical tactile sensor designed to be compact and is suitable for usage on various robotic end-effectors and multi-fingered hands. In addition, the contact region of \se is round with full 360$^\circ$ sensing clearly visible without blind spots or obscurance. While there have been some advances in round-shaped tactile sensors, their reproduction may fail due to complex and sensitive fabrication. \se sensing surface is more robust and easily interchangeable than previous designs, making the sensor more appealing. The estimated manufacturing cost for \se is 30 USD per sensor excluding a micro-controller. The main challenges in fabricating a compact and all-around tactile sensor are related to its small size and curved surface. However, we have devised a fabrication process based on in-depth experimentation so that the sensor is easily fabricated and robustly reproduced by novice users.

\subsection{General Structure}

An illustrative description of \se is given in Figure \ref{fig:intro_banner}. Similar to previous optical tactile sensors, the core of the design is a single camera. The camera is covered by a three-layered tube in the shape of a cylinder with an hemispherical end. The inner layer of the tube is a rigid crystal-clear shell. A transparent elastomer covers the shell and is coated on its exterior by a reflective silicone paint. Such tube formation provides an opaque structure where the camera observes the deformation of the elastomer from within upon contact. For better visibility, the inner-surface of the shell is evenly illuminated by an annular printed circuit board (PCB) with embedded LEDs. 
Photometric effects and structured lighting enable the camera to detect small deformations of the elastomer in physical contact. Prior work uses either white or RGB lights in different variations for contact localization and shape reconstruction. A collimator covers the LED PCB for channeling the light towards the shell and for minimizing illumination losses. All components are assembled on a mounting plate which is the connecting link to a desired hand. Unlike other sensor designs, \se is the smallest all-around tactile sensor which supports various elastomers and illumination configurations with simple assembly. Experiments in this work provide analyses to common variations.



\subsection{Fabrication}

As described above, \se has six main components: camera, mounting plate, customized LED PCB, collimator, shell and elastomer. The fabrication process for these components, illustrated in Figure \ref{fig:Manufacturing}, is described including design principles and lessons learned. 

\subsubsection*{Camera} To keep the \se sensor compact and accessible, a Raspberry-Pi zero camera is used. The camera is inexpensive costing approximately \$16 and has a wide 160$^\circ$ fisheye lens. Video is streamed directly to a PC via USB using Raspberry-Pi Zero with camera mode for easy plug-and-play support. It operates at a frame rate of 60fps and outputs $640\times480$ resolution frames. Similar to \cite{wang2021gelsight}, in order to obtain color images that are uniform and balanced, it is crucial to disable the automatic white balance function and adjust the fixed gains for the red and blue channels, along with the exposure compensation for the RGB channels.

\subsubsection*{Rigid transparent shell} The purpose of the shell is to provide rigidity to the structure of the sensor upon contact while enabling clear visibility of the external deformed elastomer. Different methods for fabricating the shell were experimented including clear epoxy resin \cite{romero2020soft}, off-the-shelf plastic tube \cite{gomes2020geltip} and a printable skeleton \cite{sun2022soft}. While the clear epoxy resin allows complex designs, the resulting shell was not sufficiently clear and required too many fabrication steps. The plastic tube is clear yet not modular. A 3D-printed skeleton provides modularity while not strong enough to withstand various pressures and point contacts within its gaps. In addition, occlusions by the ribs exist. Therefore, we propose fabrication through Stereolithography (SLA) 3D printing. The shell is designed in a custom size and shape which can be modified and scaled. Then, the shell is printed with clear resin followed by surface polishing and application of lacquer. Such approach provides both a crystal-clear shell and modularity. The shell can be easily adapted to additional shapes or scaled.

\subsubsection*{Elastomer}
The elastomer covers the entire exterior of the shell. In such way, the camera can observe deformations of the elastomer through the shell. Here also, the elastomer is made relatively clear. However, in this work we test two designs of the elastomer, both clear while one has additional dot markers. The elastomer is fabricated through molding with a two-piece mold seen in Figure \ref{fig:Manufacturing}. Different materials were tested for 3D-printed mold including SLA resin and Polylactic Acid (PLA). SLA was chosen as it provides a much smoother mold surface which affects the quality of the elastomer. For the dotted elastomer, our approach does not require any complex laser cutting \cite{taylor2022gelslim} but merely printing a mold with tiny spikes. Smooth-On Solaris\texttrademark~ is a clear and colorless silicon used for molding the elastomer. Also, it was found to be resistant to tearing and better suitable for in-hand manipulation \cite{lambeta2020digit}. Prior to casting, the interior of the mold was sprayed with lacquer to prevent sticking and allow easy release. After casting, the mold should be placed in a vacuum desiccator at a pressure of 1 bar for removing gas bubbles within the silicon. Having bubbles may damage the clarity of the sensor and prevent its robustness in model transfer. The mold is left to cure for approximately 24 hours. Finally, the elastomer is carefully removed from the mold and glued to the shell using a clear silicone adhesive (e.g., Smooth-On Sil-Poxy\texttrademark).   

\subsubsection*{Reflective coating} 
The exterior of the elastomer is coated with an opaque reflective material aimed to contain and intensify the lighting within the shell. Aluminum powder and grey silicone ink were tested for coatings while the latter proved to be more robust to wear and tear. The silicone base coating is applied by mixing an ink catalyst with a Print-On\texttrademark~ gray silicone ink and Smooth-On NOVOCS\texttrademark~ silicone solvent gloss in a 1:10:30 mass ratio. Upon testing, the coating was yet prone to tearing. Hence, we formulated a new mixture of silicone by adding Smooth-On EcoFlex\texttrademark~ (00-10) to the mixture as suggested in \cite{lambeta2020digit}. The final mixture ratio of 1:10:10:30 was used for catalyst, paint, gel and solvent, respectively. The acquired coating tested durable and reliable upon high contact forces. Prior to applying the coating for the dotted elastomer, the notches formed by the spiked mold were coated in a dip-and-wipe method. The elastomer was covered with a black silicon pigment in the same mass ratio as the reflective marker and then wiped. Only the notches retained the black paint after wiping. The reflective coating was then applied.

\subsubsection*{Mounting plate} 
The mounting plate is 3D printed with either FDM or SLA printers. Hence, \se can be adapted to various robotic hands by simply modifying the design of the interfacing plate. 

\subsubsection*{Illumination} While off-the-shelf LED PCBs are available, they can be bulky and limit the design. Hence, a customized annular PCB was designed. The PCB includes three sets of LEDs with a total of nine LEDs. Note that different combinations of LED colors are supported. Hence, we evaluate and compare between three sequences of LEDs including all white \cite{roberge2023stereotac}, RRRGGGBBB \cite{gomes2020geltip, romero2020soft} and RGBRGBRGB \cite{sun2022soft} as seen in Figure \ref{fig:intro_banner}. These combinations will be analyzed for performance. The PCB is placed between the mounting plate, while surrounding the camera, and the edge of the elastomer. The LEDs produce nine cones of light. In order to provide a uniform illumination pattern, a collimator was designed and 3D printed for light piping \cite{romero2020soft, sun2022soft}. The collimator is a ring covering the PCB with holes that adjust the direction of the lighting into the volume of the elastomer.

\begin{figure}[t]
\centering
\includegraphics[width=\linewidth,keepaspectratio]{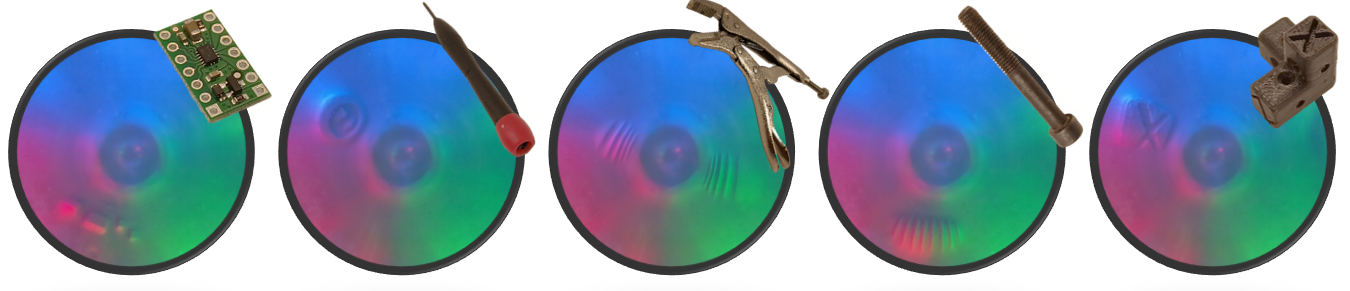} 
\vspace{-0.6cm}
\caption{Tactile images during contact of \se with (left to right) a circuit board, a screw driver, a locking pliers, a screw and an embossed 'x'.}
\vspace{-0.6cm}
\label{fig:tactile_examples}
\end{figure}

All components are assembled onto the mounting plate with three screws. The sensor is designed such that each component can easily be replaced or modified. The final shape of the assembled sensor yields a membrane with an hemisphere of 24 mm diameter on a 14 mm height cylindrical base. The mounting plate with the PCB and collimator has a height of 12 mm. Figure \ref{fig:tactile_examples} shows examples of high-resolution and clear tactile images during contact with various objects.

\section{Tactile State Learning}
\label{sec:method}

\begin{wrapfigure}{r}{0.5\linewidth}
\centering
\vspace{-0.2cm}
\includegraphics[width=\linewidth]{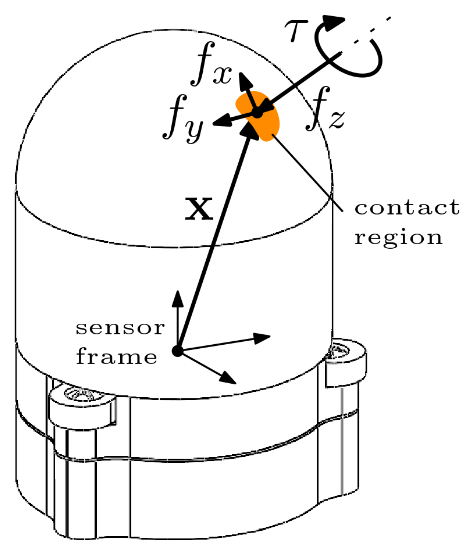} 
\caption{The contact state is defined by the position $\ve{x}$ of contact with respect to the sensors coordinate frame, normal force $f_z$ at the contact, tangential forces $f_x$ and $f_y$ and the torsion $\tau$ about the normal axis.}
\label{fig:state}
\end{wrapfigure}

A contact state $\ve{s}\in\mathbb{R}^7$ of \se is defined by the spatial location of contact $\ve{x}\in\mathbb{R}^3$ on the shell, force vector $\ve{f}\in\mathbb{R}^3$ at the contact point, and torsion $\tau\in\mathbb{R}$ with respect to the normal at the contact. Note that a force vector at the contact includes the normal force $f_z$ and tangential forces $f_x$ and $f_y$ as seen in Figure \ref{fig:state}. While can also be included in the state, contact torques about the $x$ and $y$ axes are disregarded as they are insignificant to general in-hand manipulation tasks. It is important to note that, in this work, we evaluate models for point contacts as often occur in in-hand manipulations of various objects. However, the sensor can be used with more complex contacts such as in the manipulation of soft objects. The proposed approach for training a state estimation model based on real and simulated image datasets is illustrated in Figure \ref{fig:sketch} and discussed next.
\vspace{-0.2cm}

\subsection{Data collection}
\vspace{-0.1cm}

We use two sources of training data:
\subsubsection{Real-world data} Dataset $\mathcal{P}_{real}$ is collected by labeling images captured by the internal camera during premeditated contact. A robotic arm equipped with a Force/Torque (F/T) sensor and an indenter touch the surface of the sensor in various contact locations and loads. During contact, an image $\ve{I}_i$ is taken along with a state measurement $\ve{s}_i$. Contact position  $\ve{x}_i$ is calculated through the forward kinematics of the arm. Load at the contact (i.e., force vector $\ve{f}_i$ and torsion $\tau_i$) is measured by the F/T sensor fixed at the wrist of the robotic arm. In addition to the contact state, the maximum penetration depth $d_i$ of the indenter is also measured. 
The acquisition and labeling process yields dataset $\mathcal{P}_{real}=\{(\ve{I}_i,\ve{x}_i,\ve{f}_i,\tau_i,d_i) \}_{i=1}^N$ of $N$ labeled images. In addition, reference image $\ve{I}_{ref}$ is recorded for a sensor without any contact.
\begin{figure*}[ht]
\centering
\includegraphics[width=\textwidth]{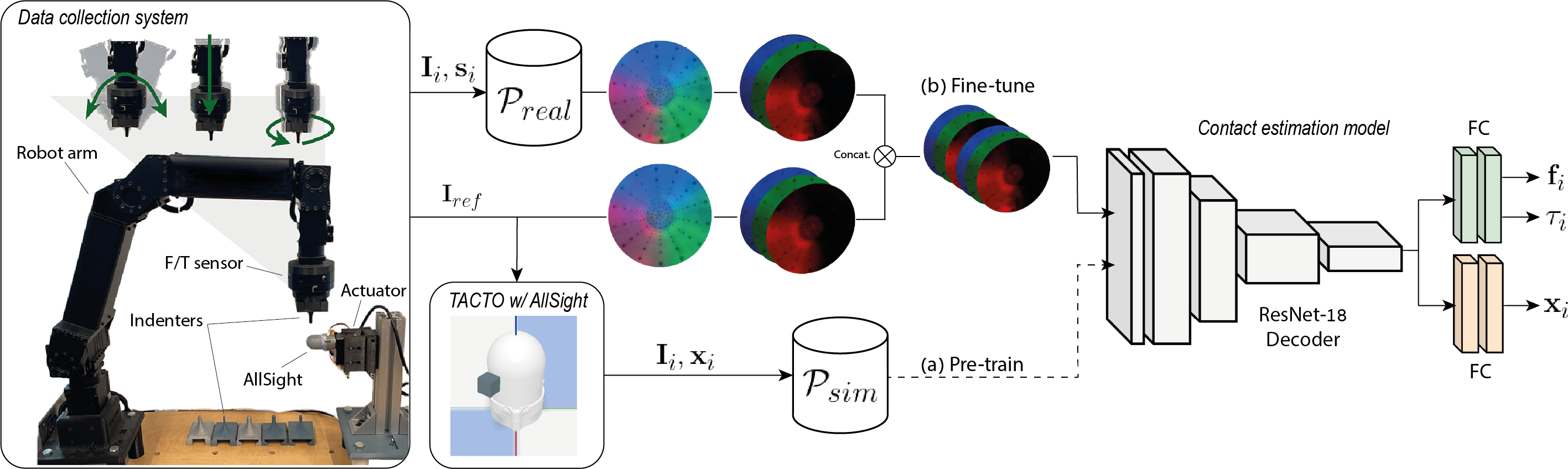} 
\caption{Data collection system including a robotic arm and an F/T sensor with several indenters. During premeditated contact, images are labeled with contact location and loads. (a) The contact localization part of the model is pre-trained using simulated data from TACTO. (b) Then, the entire state estimation model is fine-tuned using real data. Reference images from the real sensor are used for augmenting the collected data and for generating simulated data for pre-training the contact model.}
\label{fig:sketch}
\vspace{-0.5cm}
\end{figure*}

\vspace{-0.3cm}
\subsubsection{Simulated data} \se was implemented in the TACTO physics-engine simulator for optical-based tactile sensors \cite{wang2022tacto}. In TACTO, we calibrated the renderer to sufficiently match the real-world by including reference images from real \se sensors. To enable and optimize sim-to-real pre-training of the state estimation model, we collected different reference images from different \se sensors and used them for augmentation. The acquired images were augmented by adding noise and varying the lighting conditions. TACTO simulator does not support marker motion and, therefore, only images from \se sensors with clear shells were used. 
A simulated dataset $\mathcal{P}_{sim}$ was generated by labeling $M$ images captured in TACTO during random premeditated contacts. During contact, an image $\ve{I}_i$ is taken along with the contact position $\ve{x}_i$ such that $\mathcal{P}_{sim}=\{(\ve{I}_i,\ve{x}_i) \}_{i=1}^M$. Penetration depth $d_i$ can also be acquired but not used here.

\vspace{-0.3cm}
\subsection{State estimation}
\label{sec:state_est}

\vspace{-0.1cm}
We adopt a modified ResNet-18 architecture \cite{he2016deep} as the state estimation model. The top layer is removed and the flattened output features are fed through two Fully-Connected (FC) layers of size 512 and 256, and with ReLU activation functions. At each iteration, both reference $\ve{I}_{ref}$ and contact $\ve{I}_t$ images are down-sampled to resolution $224\times224$ and stacked along the channel. The stacked image is then passed through the model to get the estimated state $\tve{s}_t$. Simulation data offers a means to collect training data with a much lower effort. While a simulator often cannot provide data similar to the real world, one can pre-train a model prior to fine-tuning it with real data. In such way, a smaller dataset of real world data is required. Hence, the decoder of the contact localization model to approximate $\ve{x}$ is pre-trained with $\mathcal{P}_{sim}$. Finally, the entire contact model is fine-tuned on the real dataset $\mathcal{P}_{real}$. 




\section{Experiments}
\label{sec:experiments}

\subsection{Data collection}

\textbf{Simulated dataset.} Dataset $\mathcal{P}_{sim}$ comprises of simulated tactile images and corresponding contact poses involving six types of indenters: three spherical indenters, one rectangular, one elliptical and one squared. These indenters were utilized only on \se with clear shells. To calibrate the simulation, we employed reference images from six different sensors. In addition, Gaussian noise and various illumination settings were used to augment the simulated images. These are intended to make the model independent of the background and focus on capturing the color gradient observed at the contact pixels. For the localization pre-training, our dataset consists of $18,000$ samples, with $1,500$ samples allocated for each indenter-configuration pair.


\textbf{Real dataset.}
As described in Section \ref{sec:state_est}, dataset $\mathcal{P}_{real}$ is collected in an automated process. The collection setup, seen in Figure \ref{fig:sketch}. An indenter is mounted to an OpenMANIPULATOR-P arm equipped with a Robotiq FT-300 \ft sensor. An AllSight sensor is fixed on the axis of an actuator in order to increase the reachability of the arm. The system is controlled using the Robot Operating System (ROS). During the collection, data stream is acquired in a frequency of 60 Hz. The train and test data are collected in episodes where, at each episode, the robot selects a contact point to press on the sensor's surface. Upon contact, the arm either presses perpendicular to the surface, tilts back and forth about the normal to the surface in order to exert tangential forces or twists the end effector with respect to surface normal for torsion samples. These are chosen arbitrarily and in varying magnitudes within the ranges $f_z\in[-12N,0.8N]$, $f_x,f_y\in[-5N,5N]$ and $\tau\in[-0.05Nm,0.05Nm]$. During the pressing, images are taken in $480\times480$ resolution, after circular masking and centering, along with contact states. 
3D-printed indenters are used for generating different contact geometries including round indenters of radius 3, 4 and 5 mm, square (edge length 6~mm), hexagonal (edge length 3~mm) and elliptical (axis lengths 8~mm and 4~mm) heads. 

\subsection{Contact State Estimation}
\vspace{-0.1cm}

We assess the precision of state estimation using collected data. To compare performance, we evaluate a series of six sensor configurations seen in Figure \ref{fig:intro_banner}. These configurations involve cross combination of shells with and without markers, along with three illumination setups: all white, RRRGGGBBB, and RGBRGBRGB. Several experiments were conducted to evaluate the contact estimation capabilities of \sep. In each experiment, the dataset was divided, allocating 80\% for training and 20\% for testing purposes.

\begin{figure}[h]
\centering
\vspace{-0.3cm}
\includegraphics[width=\linewidth]{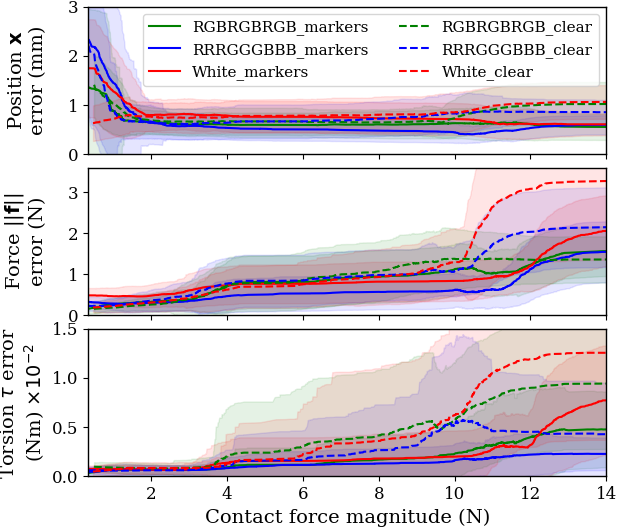} 
\vspace{-0.6cm}
\caption{State estimation errors for the six \se configurations with respect to the  force magnitude $\|\ve{f}\|$ at the contact.}
\label{fig:eval}
\vspace{-0.2cm}
\end{figure}

In the first experiment, a comparative analysis was conducted among the six different AllSight configurations. Each sensor was trained using optimized hyper-parameters, utilizing $N=12,000$ collected samples in $\mathcal{P}_{real}$ featuring a single spherical indenter of 3 mm radius. 
Figure \ref{fig:eval} shows state estimation results over the test set with three different spherical indenters, 3~mm, 4~mm and 5~mm radii. The contact location, force magnitude $\|\ve{f}\|$ (with $\ve{f}=(f_x,f_y,f_z)^T$) and torsion errors are shown with with respect to $\|\ve{f}\|$. Results show that all configurations achieve low estimation errors with subtle differences. Nevertheless, using markers provide lower errors compared to clear elastomers. Overall, the RRRGGGBBB with markers configuration provides the best estimation. Nevertheless, the differences are neglectable for moderate forces. Markers were often reported to interfere with depth estimation and 3D reconstruction \cite{wang2021gelsight}. To evaluate this, we also separately trained models to estimate the penetration depth $d_i$ of the indenter. Low mean errors of 0.15$\pm$0.14 mm and 0.19$\pm$0.16 mm with and without markers, respectively, were obtained. The markers are positioned rather sparse and maintain the ability to observe gradients and estimate depth. Hence, one can choose to use an elastomer with or without markers, with minimal trade-off. In general, the accurate depth estimation indicates the sensor's ability to provide reliable geometric information about the contact shape.

\begin{figure}[h]
\centering
\vspace{-0.3cm}
\begin{tabular}{ccc}
    \hspace{-0.4cm}
    \includegraphics[width=0.32\linewidth]{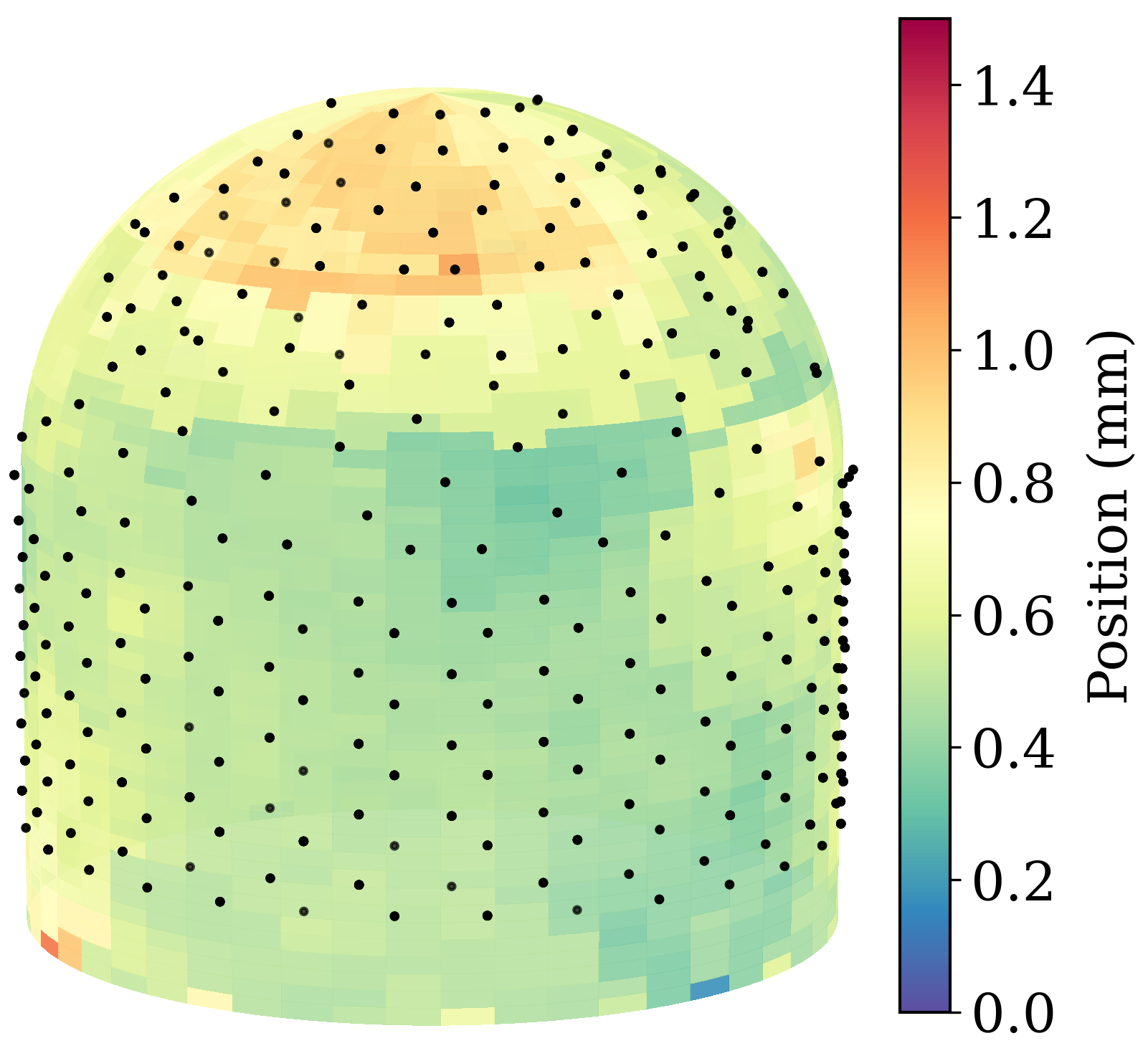}  & \hspace{-0.3cm} \includegraphics[width=0.32\linewidth]{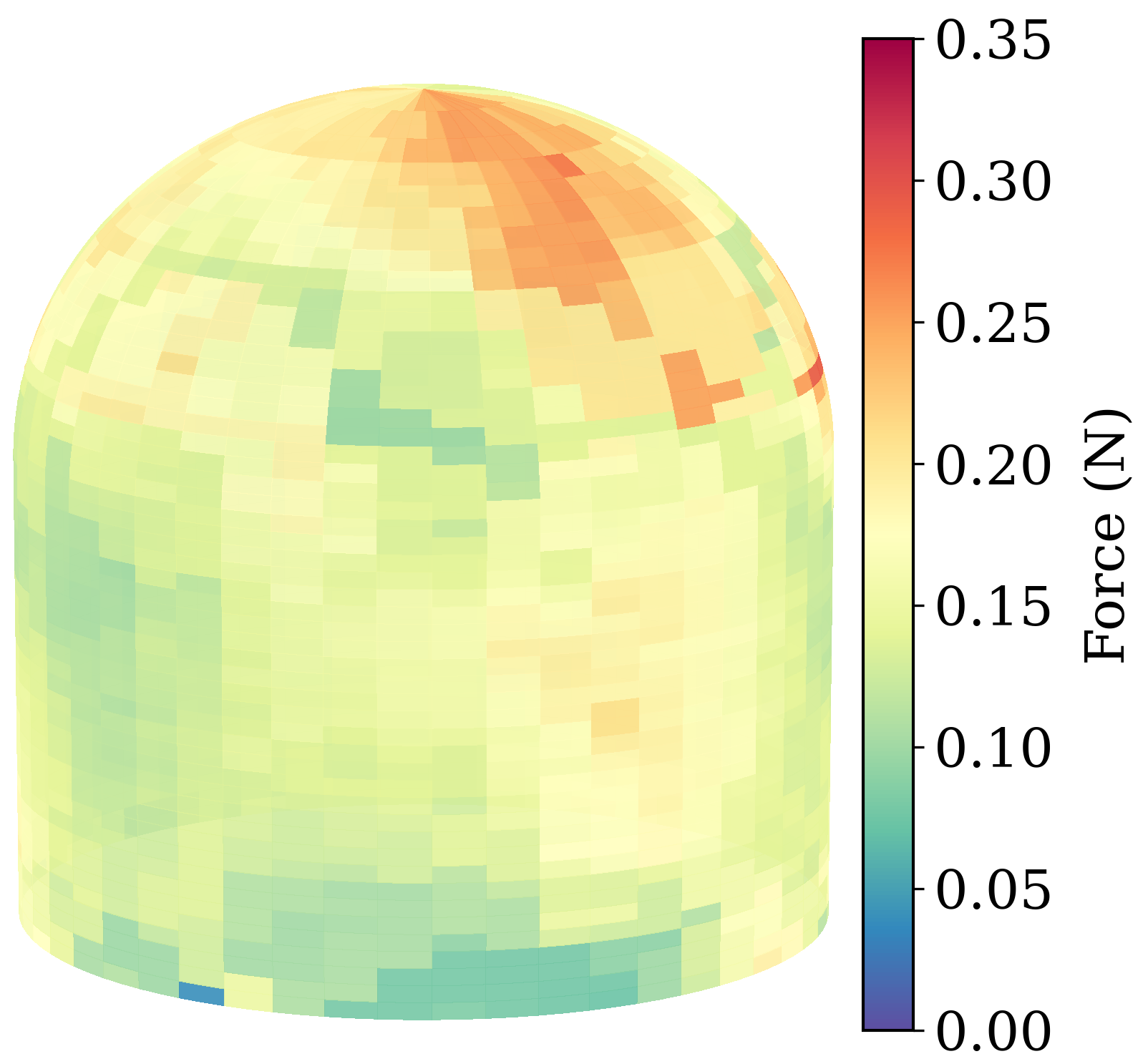} & \hspace{-0.3cm}\includegraphics[width=0.32\linewidth]{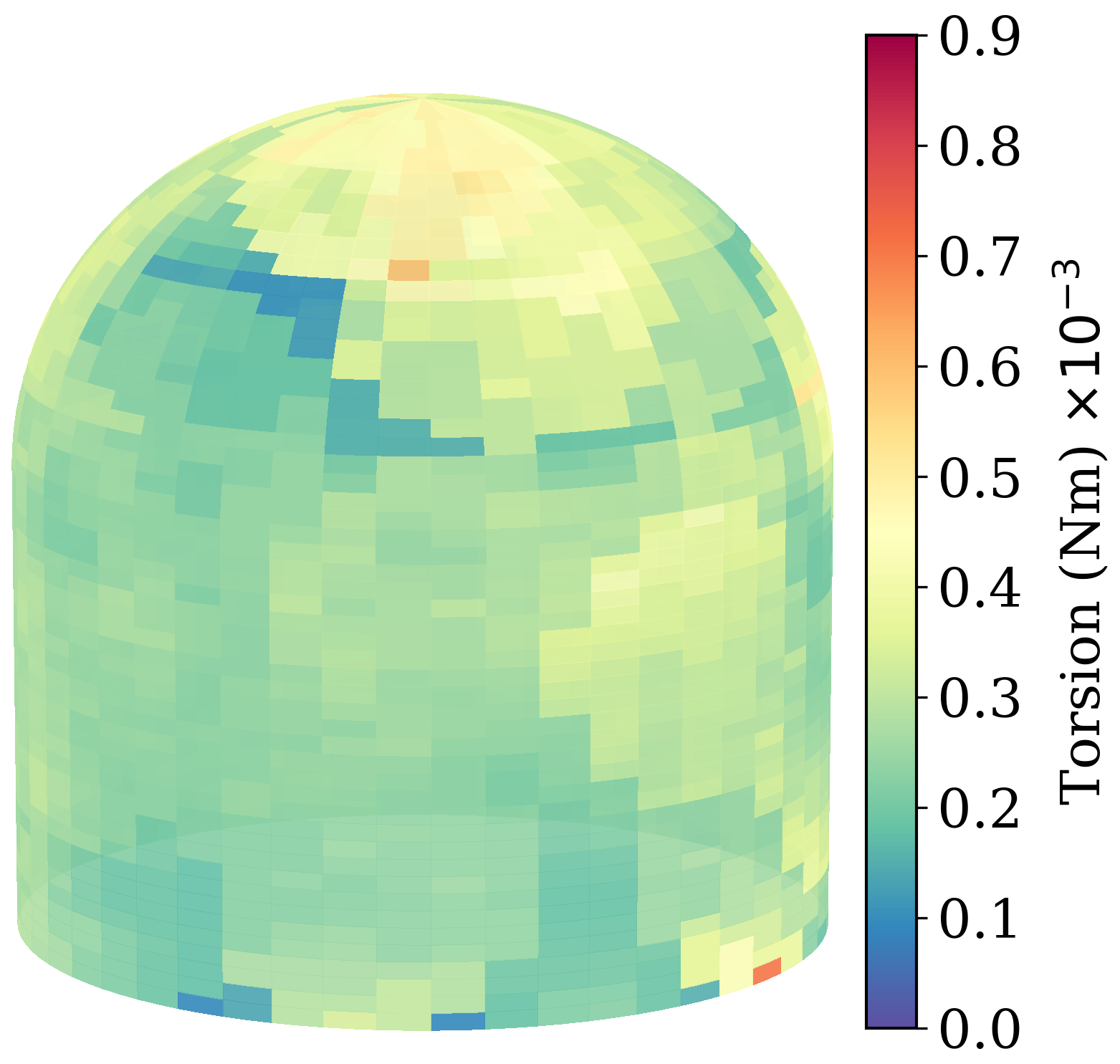}  \\
    (a) & (b) & (c) \\
\end{tabular}
\caption{Heatmap of (a) position with markers indicating the test samples, (b) force and (c) torsion estimation errors on test data from different indenters and with respect to contact position.}
\label{fig:eval_diff_ind}
\vspace{-0.5cm}
\end{figure}

The above model was trained on data collected solely from spherical indenters. Hence, we observe the ability to learn a model for contact state estimation with various indenter geometries. A model was trained for the  RRRGGGBBB-markers configuration with a data of size $N=20,000$ samples featuring spherical, hexagonal, ellipse and square headed indenters. Figure \ref{fig:eval_diff_ind} illustrates a heatmap of state estimation errors with respect to the indenter position on the contact surface evaluated over 1282 test points equally distributed on the surface. The mean position, force and torsion errors are significantly low at 0.59 mm, 0.15 N and 0.0002 Nm, respectively. 
Videos of experiments and demonstrations with the sensor, including in-hand manipulation (Figure \ref{fig:intro}), can be seen in the supplementary material.

\subsection{Data Efficiency}

\begin{figure}[ht!]
\centering
\includegraphics[width=\linewidth]{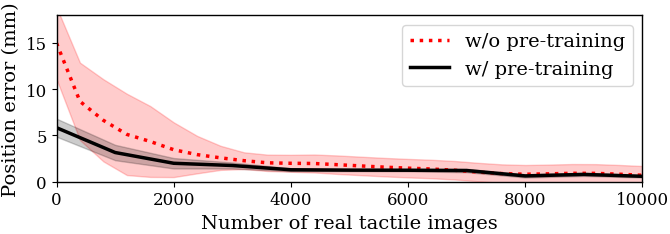} 
\vspace{-0.6cm}
\caption{Position estimation accuracy with regards to the amount of real training data with and without pre-training using simulated data. The curves represent mean and standard deviation errors over the test data.}
\label{fig:sim2real}
\vspace{-0.6cm}
\end{figure}

The above trained models were based on pre-training of the model with simulated data. We now evaluate the contribution of the simulation and the sim-to-real effort in the fine-tuning of the model. First, $1,000$ samples from a real RRRGGGBBB-Clear sensor were taken as test data for evaluation. Two models were trained with up to $10,000$ real samples: one without any fine-tuning while the second was initially pre-trained with $4,500$ simulated samples. Figure \ref{fig:sim2real} shows the accuracy improvement over the test data for both models with the addition of real training samples. In zero-shot and no fine-tuning, the pre-trained model already provides good performance. With a small amount of real data (approximately 2,000 samples) for fine-tuning, the pre-trained model achieves good performance. These results emphasize the contribution of the simulation for reducing the effort of real data collection. 


\subsection{Transfer Learning}

The ability of the proposed \se sensor to provide an accurate contact state estimation has been shown above. Nevertheless, the analysis was performed on the same sensor that was used to collect training data. A practitioner may desire to instantly use a newly fabricated sensor without further training, i.e., zero-shot inference. Alternatively, the practitioner can collect a limited amount of labeled data from the new sensor and fine-tune the model for better results. Hence, we now wish to observe the ability of transferring a learned state estimation model to a newly fabricated sensor. Expanding upon previous findings that demonstrated superior results of the RRRGGGBBB-marker configuration, we conducted an assessment of transfer learning capabilities for the \se sensor on both zero-shot and fine-tuning approaches. 
\begin{figure}[ht!]
\centering
\includegraphics[width=\linewidth]{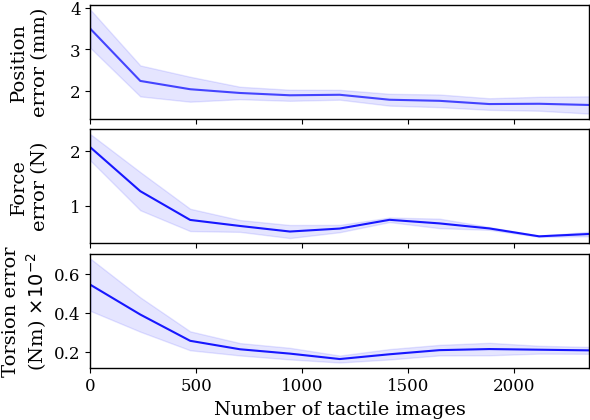} 
\vspace{-0.7cm}
\caption{Transfer estimation errors for a newly fabricated \se (RRRGGGBBB-markers) with regards to the number of new samples used for fine-tuning the model. Results with zero new tactile images are the zero-shot transfer errors without any fine-tuning.}
\label{fig:transfer}
\vspace{-0.3cm}
\end{figure}

We strengthen the state estimation model and train it with $N=40,000$ samples collected from three distinct sensors using round indenters of 3~mm, 4~mm and 5~mm radii. To enhance the ability of the model to generalize, we augmented the images with lighting randomization. Furthermore, $2,300$ image samples labeled with full contact states were collected from a newly fabricated sensor for possible fine-tuning of the model. An additional $600$ labeled image samples not included in the training were collected from the new sensor for testing the model. 

Figure \ref{fig:transfer} exhibits the accuracy of state estimation over the test data of the new sensor with regards to the number of new samples used to fine-tune the model. First, when no data was used to fine-tune the model, i.e., zero-shot inference, the position, force and torsion errors are already low at 3.49$\pm$0.41~mm, 2.06$\pm$0.23~N and 0.0068$\pm$0.0016~Nm, respectively. With the addition of a limited amount of new training data for fine-tuning, accuracy was improved even more. The results provide compelling evidence that the \se sensor achieves satisfactory zero-shot performance with ability to further improve. We attribute this success to the robust design of the \se sensor, the incorporation of reference images and the benefits gained from training data of multiple sensors. Evidently, a practitioner can fabricate our open-source \se sensor and have a ready-to-go operational model.

\section{Conclusions}
In this paper, we have proposed \sep, an optical tactile sensor with a round surface. The fabrication process was discussed where most of the components including the clear rigid shell are 3D printed. The design is fully open-source with fabrication instructions for the benefit of the robotics community. Furthermore, we have presented a comprehensive analysis for contact state estimation accuracy provided by possible design choices. Results show that RRRGGGBBB illumination with dotted markers on the elastomer provide best results by a small margin. Furthermore, we show the advantageous use of simulated data to pre-train a state estimation model in order to reduce the amount of real data to collect. With the proposed design and training process, \se is shown to have a unique zero-shot learning capability where a newly fabricated sensor can use a model trained on other sensors with sufficient accuracy. Therefore, \se is a low-cost, small, modular and open-source sensor with a ready-to-go state estimation model. Limitations which should be addressed in future work are the augmentation of the simulation with real-like tactile images and force perception in order to ease the sim-to-real transfer. In terms of design, one can develop a new LED PCB that can fit various radii of the shell for increased modularity. The next step in future work can be to push the boundaries of size and additionally minimize the sensor for smaller manipulation tasks. In addition, multi-\sep-sensor inference should be analyzed for precise control of objects in manipulation tasks.

\bibliographystyle{IEEEtran}
\bibliography{ref}

\end{document}